%% file: main.tex
\documentclass[twoside,journal]{IEEEtran}
\pdfoutput=1
\usepackage{color}
\usepackage{amsmath}
\usepackage{amssymb}
\usepackage{graphicx}
\usepackage{algorithm}
\usepackage{algorithmic}
\usepackage{enumitem}
\usepackage{bm}
\usepackage{microtype}
\usepackage{threeparttable}
\ifCLASSOPTIONcompsoc
    \usepackage[caption=false, font=normalsize, labelfont=sf, textfont=sf]{subfig}
\else
    \usepackage[caption=false, font=footnotesize]{subfig}
\fi

\usepackage{url}
\newcommand{\TitleLong}{Robust Imitation Learning for Automated Game Testing}
\newcommand{\TitleShort}{Imitation Learning for Automated Game Testing}
\newcommand{\proposed}{EVOLUTE}
\newcommand{\ebmlane}{EBM-DNN}
\newcommand{\bclane}{FF-DNN}
\newcommand{\purebc}{FF-BC}

\begin{document}
\bstctlcite{IEEEexample:BSTcontrol}

\title{\TitleLong}

\author{Pierluigi~Vito~Amadori,~\IEEEmembership{Member,~IEEE},
        Timothy Bradley,
        Ryan Spick,
        Guy~Moss

        % <-this % stops a space

%\thanks{Manuscript submitted \today.}
\thanks{All authors are with the Future Technology Group, Sony Interactive Entertainment Europe, London, UK.
~(e-mail:~{\tt\footnotesize \{pierluigi.vito.amadori, timothy.bradley, ryan.spick, guy.moss \}@sony.com})}% <-this % stops a space
% \thanks{}% <-this % stops a space
}

\markboth{Amadori \MakeLowercase{\textit{et al.}}: \TitleShort}
{Amadori \MakeLowercase{\textit{et al.}}: \TitleShort}

\maketitle

\input{sections/0_abstract.tex}


\IEEEpeerreviewmaketitle

\input{sections/1_introduction}

\input{sections/2_related_works}
\input{sections/3_model}
\input{sections/4_ebms}
\input{sections/5_results}
\input{sections/6_conclusions}

\bibliography{bibliography}
\bibliographystyle{IEEEtran}

\end{document}

%% file: sections/0_abstract.tex
\begin{abstract}
Game development is a long process that involves many stages before a product is ready for the market. Human play testing is among the most time consuming, as testers are required to repeatedly perform tasks in the search for errors in the code. Therefore, automated testing is seen as a key technology for the gaming industry, as it would dramatically improve development costs and efficiency. Toward this end, we propose \proposed{}, a novel imitation learning-based architecture that combines behavioural cloning (BC) with energy based models (EBMs).
\proposed{} is a two-stream ensemble model that splits the action space of autonomous agents into continuous and discrete tasks. The EBM stream handles the continuous tasks, to have a more refined and adaptive control, while the BC stream handles discrete actions, to ease training. We evaluate the performance of \proposed{} in a shooting-and-driving game, where the agent is required to navigate and continuously identify targets to attack.
The proposed model has higher generalisation capabilities than standard BC approaches, showing a wider range of behaviours and higher performances. Also, \proposed{} is easier to train than a pure end-to-end EBM model, as discrete tasks can be quite sparse in the dataset and cause model training to explore a much wider set of possible actions while training.

\end{abstract}

%% file: sections/1_introduction.tex
\section{Introduction}
\IEEEPARstart{D}{eveloping} games is a very complex process that requires many iterations of human testing to evaluate the quality of the experience for the player. Testing can help identify errors in the game behaviour and, more generally, is used to ensure the quality of the product. While fundamental to game development, human play testing is very costly and is not a scalable solution, as it requires human participants to test the game over many hours~\cite{Politowski2021}. The process is also very demanding on the participants themselves, as they need to perform repetitive and unengaging tasks several times over. Because of this, the gaming industry has looked for ways to automate and streamline this process. Recent advances in reinforcement learning (RL) and imitation learning (IL) have fuelled the interest in applying these technologies to automated testing. While autonomous agents in RL show impressive performance~\cite{Wurman2022}, our focus resides on IL as it does not require specific integration with the game in and can operate with readily available human demonstrations.

Imitation learning or learning from demonstration is a powerful yet simple approach to train autonomous agents or robots to perform new tasks. At its core, IL assumes that autonomous agents can learn the appropriate controls to perform a task from an offline dataset that consists of demonstrations (or trajectories) of a human expert performing the task. Despite its simplicity, IL has been used to train agents to perform many tasks over the years, ranging from robotics~\cite{Schaal1999, Johns2021, Mu2021} to self-driving cars~\cite{Bojarski2016, Bansal2018}. 

The simplest approach to IL is Behavioural Cloning (BC), which formalises policy learning as a supervised learning problem. Here, state observations are used as input to the model and actions identify the labels or classes for the model. Due to its simplicity, BC is normally used as a first step to train autonomous agents as it allows to quickly teach basic control behaviours~\cite{Pomerleau1991, Bojarski2016}. However, this ease of implementation is accompanied by known shortcomings, such as inability to generalise, compounding errors and trajectory distribution drifts~\cite{Ross2011, Laskey2017}.

Ingenious learning paradigms and complex dataset handling~\cite{Ross2011, Laskey2017} have helped boost BC performance over the years, but they all rely on the same feed-forward (FF) neural network model architecture for the policy model. While FF models are very powerful and can lead to great results on a multitude of tasks, they are not suited to model the uncertainty that often characterises behavioural policies. To tackle this, recent studies~\cite{Florence2022} have proposed the use of mixed density networks and energy-based models (EBMs)~\cite{Lecun2006} when training policies in a BC framework. EBMs in particular have experienced a surging interest in recent years due to their high generalisation capabilities in regression~\cite{Gustafsson2020}, classification~\cite{Grathwohl2019} and generative problems~\cite{Zhao2016}.

\begin{figure}[t]
  \centering
  \vspace{0.15cm}
  \includegraphics[width=0.99\linewidth,trim=4cm 2cm 4cm 1.8cm,clip]{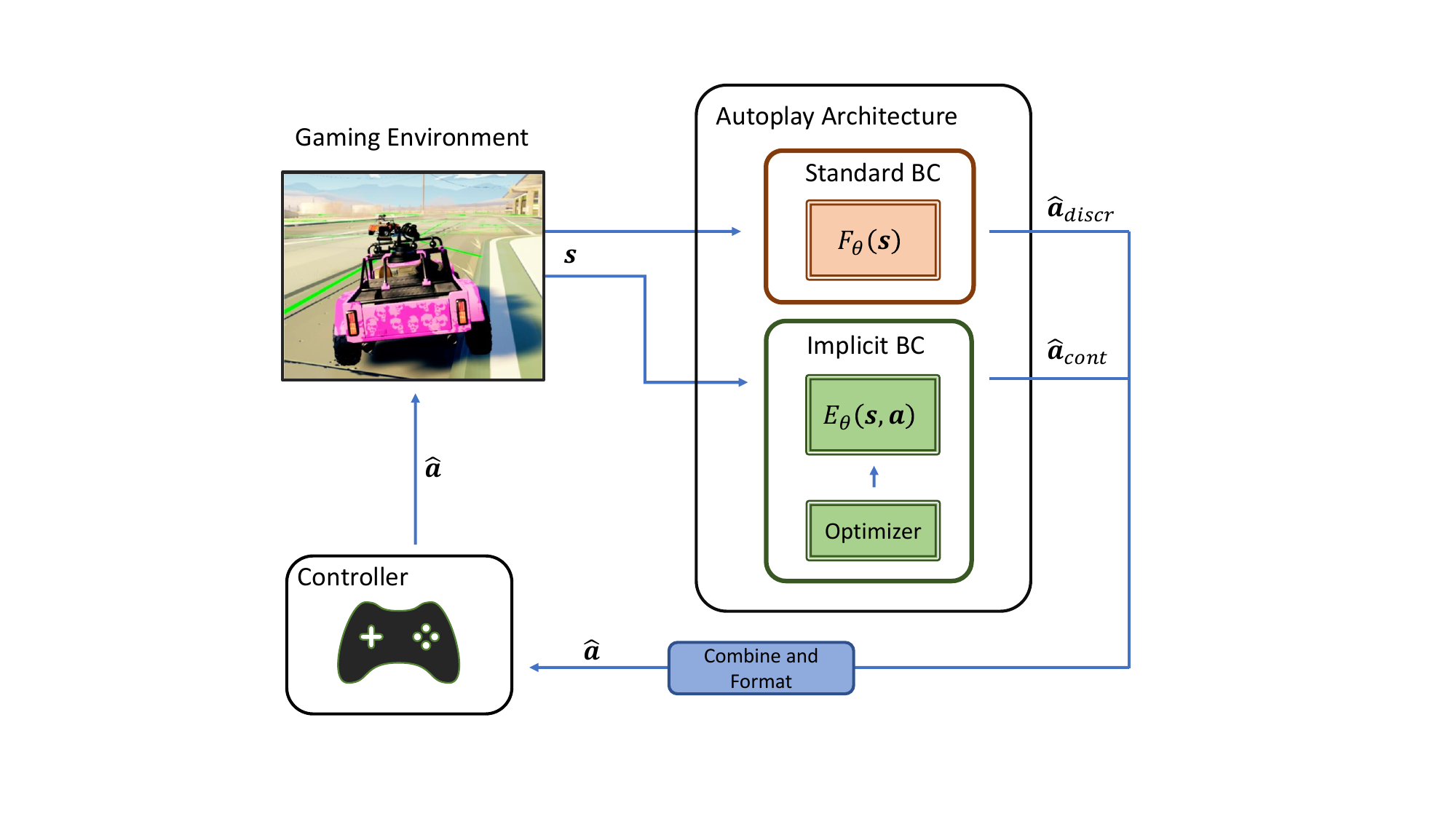}%
  \caption{Diagram of \proposed{}. Standard FF-based BC (brown) handles discrete actions, while EBM-based BC (green) controls continuous actions. Once the estimates of the two set of actions are computed, they are combined (blue) and sent to the gaming environment.}%
  \label{fig:fig1}
  \vspace{-0.25cm}
\end{figure}

In the studies above, videogames have been often used as a testing platform for policy learning. Videogames and virtual environments are intrinsically episodic, offer full control and repeatability of the environment~\cite{Mnih2013}. However, the final goal of these methods is often aimed at robotics or intelligent transportation systems applications. In this paper, we propose a novel ensemble model whose goal is to reliably and robustly imitate the behaviour of a human player while exploring the environment to automate game testing. In this environment, the trade-off between ease of implementation,  model training complexity and policy performance plays a delicate role, as more high-performing but brittle policies would require costly iterations and demanding tool integration within game development.

The contribution of the paper are threefold, as follows:
\begin{enumerate}[topsep=0pt,itemsep=2pt]
\item We introduce g\textbf{E}nerati\textbf{V}e p\textbf{O}wered ensemb\textbf{L}e for a\textbf{UT}onomous gam\textbf{E}play (\proposed{}), a model that combines in an end-to-end fashion standard BC and energy controlled BC;
\item We evaluate \proposed{} performance over two separate gaming scenarios and compare it against comparable baselines. Our results indicate that the proposed architecture can achieve $2$ kills on average and consistently stay alive for the whole duration of the match;
\item We investigate how well \proposed{} can generalise to unseen situations when deployed in gaming scenarios and evaluate its ability to explore the environment, despite being trained on little data.
\end{enumerate}

Following this introductory section, Section~\ref{sect:related} presents related literature on imitation learning. We introduce the proposed architecture in Section~\ref{sect:model}, together with a detailed formalisation of the problem and of energy-based models. In Section~\ref{sect:ebms}, we describe in more depth energy-based models, providing a detailed description of training and inference procedures. We compare the proposed model to other baseline approaches in Section~\ref{sect:results}, highlighting the benefits \proposed{} provides in terms of robustness, generalisability and adaptability. Finally, we summarise the contributions of the paper in Section~\ref{sect:conclusions}.

%% file: sections/2_related_works.tex
\section{Related Works}
\label{sect:related}
Our work is closely related to imitation learning, behavioural cloning in particular, and autonomous agents in games.

Behavioural cloning (BC) has experienced a continuous interest in the research community thanks to its ease of implementation. However, its simplicity comes at the cost of known shortcomings which many studies have tried to address and tackle over the years. For instance,~\cite{Ross2011, Laskey2017} have been proposed to solve what is commonly known as covariate shift in BC. Intuitively, when a learned policy makes a mistake and deviates from the expert policy, the agent is bound to experience more and more unseen situations, with errors building on top of each other often with catastrophic effects. To solve this, DAgger~\cite{Ross2011} proposes an online learning approach where the expert expands the dataset by providing correcting actions while the learned policy is deployed. In a similar way DART~\cite{Laskey2017} proposes to inject noise in the demonstrations provided by the expert, therefore allowing for the learned policy to have some robustness to small deviations in the policy.
Over the years, there have been numerous approaches to IL that go beyond BC and attempted to bridge the gap between RL and IL. Among these, we highlight Inverse Reinforcement Learning (IRL)~\cite{Ng2000, Ho2016, Wang2019}, where we first use trajectories in an offline dataset to estimate a cost function and we then use this cost function to train a policy using RL algorithms. To do so,~\cite{Ho2016} employed the power of generative adversarial networks to frame imitation learning as distribution matching, while~\cite{Wang2019} proposed to use expert trajectories to estimate a robust and fixed reward function.

While the above methods often use games as a platform to test algorithms and autonomous agents due to their ease of implementation for episodic studies, they often use autonomous agents in games as a tool or research platform with goals that fall outside the gaming industry. Other works, however, have focused on the intricacies of implementing realistic and believable autonomous agents in games so that they can actually be deployed in future games~\cite{Wurman2022}, be used as benchmark against human players~\cite{Silver2017, Vinyals2019} or be employed to support the gaming industry~\cite{Ariyurek2019}. For instance,~\cite{Wurman2022} trained high-performing RL-trained autonomous agents that can compete with professional players in a racing game and later deployed those agents in the game. On the other hand,~\cite{Silver2017} trained an agent to perform at a professional level on the complex game of GO, beating several professional players. Similarly,~\cite{Vinyals2019} famously developed a multi-agent RL algorithm that succeeded in competing professionally on Starcraft II. Finally,~\cite{Ariyurek2019} combined RL and Montecarlo tree search algorithm to train autonomous agents that look for exploits that can break a game. 

%% file: sections/3_model.tex
\section{Proposed Architecture}
\label{sect:model}

\begin{figure*}[t]
  \centering
  \vspace{0.15cm}
  \includegraphics[width=0.99\linewidth,trim=1cm 8cm 1cm 10cm,clip]{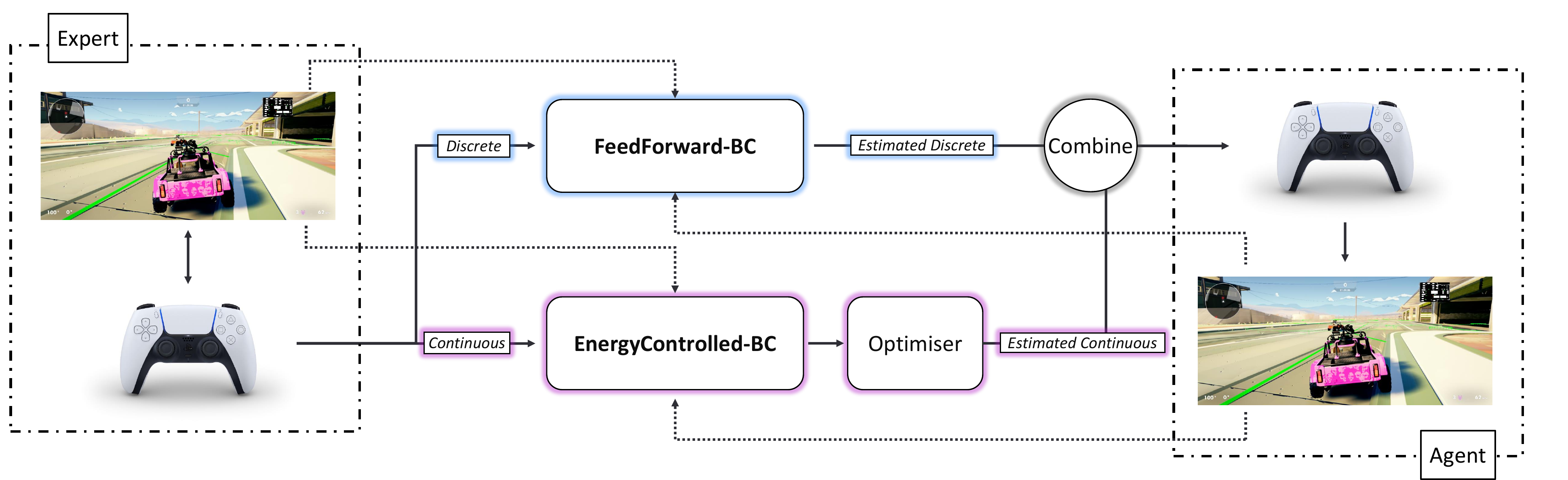}%
  \caption{Diagram of \proposed{}. }%
  \label{fig:fig2}
\end{figure*}

We propose \proposed{}, a novel two-stream ensemble model for robust IL in automated game testing. In the proposed framework, see Fig.~\ref{fig:fig2}, one stream performs standard BC with a FF-based classifier, namely FeedForward-BC (\purebc{}), while the other performs BC with an energy-based model (EBM), namely EnergyControlled-BC (EC-BC). To do so, we split the action space of the agent, and its policy, between discrete and continuous actions. Specifically, \purebc{} is dedicated to controlling the discrete actions, while EC-BC controls the continuous actions. 
The rationale is that discrete actions often identify on/off tasks or grab/non-grab tasks, which are not modeled well by probability distributions~\cite{Lillicrap2015} and can therefore be easily learned using standard FF-based BC. On the other hand, continuous tasks represent movement actions or camera controls, which are better modeled by probability distributions~\cite{Lillicrap2015, Cai2023} and are therefore better controlled by EBMs.
This assumption is particularly true in the case of autonomous agents which play video games~\cite{Delalleau2019}, the main focus of this paper. In fact, in gaming, discrete tasks often control button-related actions which can be easily formalised as binary actions, e.g., pick-up/drop item or interact/no interact. On the other hand, continuous tasks are dedicated to navigation, e.g., avatar movement and camera control.

\subsection{Problem Formulation}
Formally, we present the task as a Markov Decision Process (MDP) defined by the 4-tuple $\langle\mathcal{S}, \mathcal{A}_{c}, \mathcal{A}_{d}, p\rangle$. Here, $\mathcal{S}$ indicates the state information space, $\mathcal{A}_{c}$ the action space for continuous control inputs, $\mathcal{A}_{d}$ the action space for discrete action inputs and $p$ is the state transition distribution. We represent the dynamics of the system using the following transition model:
\begin{equation}
    p(\mathbf{s}_{t} | \mathbf{s}_{t-1}, \mathbf{a}_{t-1} )_,
\end{equation}
where $p$ is the conditional probability to reach the state $\mathbf{s}_{t}$ given the previous state $\mathbf{s}_{t-1}$ and control action $\mathbf{a}_{t-1}$.

Under these assumptions, the goal of IL is to identify a policy function $\pi (\mathbf{s})$ that associates an action $\mathbf{a}$ to the state information $\mathbf{s}$, as follows:
\begin{equation}
    \mathbf{a}_{t} = \pi (\mathbf{s}_{t}).
\end{equation}

In order to train the agent to perform a task with IL, we need access to a dataset that comprises of a set of demonstrations or trajectories of an expert performing the same task.  Such a dataset is often characterised by a sequence of tuples of state information and actions, as follows:
\begin{equation}
    \mathcal{D} = \left\lbrace (\mathbf{s}_{t}, \mathbf{a}_{t})_{k},~\forall~t, k  \right\rbrace _,
\end{equation}
where sub-indices $t \in \left[0, T\right]$ and $k \in \left[0, K \right]$, identify the time in the demonstration trajectory (of maximum duration $T$) and the trajectory (among all $K$ trajectories), respectively.

The goal of IL is to use these demonstrations to train a machine so that is can learn a policy function $\hat{\pi} (\cdot)$ that approximates the behaviour of the expert, i.e., the policy $\pi(\cdot)$. To achieve this, we indicate two separate key approaches: Behavioural Cloning (BC), where we leverage a supervised learning paradigm to approximate the policy of the expert during the demonstrations, and Inverse Reinforcement Learning (IRL), where we use trajectories to learn the reward function used by the expert and therefore indirectly imitate their policy.  While IRL has shown impressive results over the years~\cite{Wang2019, Ho2016}, its need for reward information and for online interaction of the learned policy with the environment requires a level of integration that reduces the ease of application to automated game testing, the focus of our paper.

With BC, we approximate the policy function $\pi(\cdot)$ with $\hat{\pi} (\cdot)$ within a supervised learning framework. To do so, we minimise the difference between the policy of the demonstrations and the learning policy according to a chosen metric $\mathcal{L}$. This is formally represented as the following optimisation problem:
\begin{equation}
    \mathcal{P}: \hat{\pi} = \arg \min_{\pi} \sum_{k \in K} \sum_{t \in T} \mathcal{L}(\pi(\mathbf{s}), \pi^{*}(\mathbf{s})) _,
\label{eq:opt_prob}
\end{equation}
where $\mathcal{L} (\cdot, \cdot)$ is the chosen loss metric, e.g., MSE, KL-divergence or cross-entropy. Please note that we have omitted time and trajectory sub-indices on the state vector $\mathbf{s}$ to ease notation.

When solving the optimization problem in Eq.~(\ref{eq:opt_prob}), a common approach is to have a single FF-based network for the complete policy $\pi(\cdot)$~\cite{Ross2011, Laskey2017}. While this is a viable option that has seen a certain degree of success over many problems, a single FF network fails to model the uncertainty that we might experience when deploying policies in complex environments. For instance, an expert driving policy might steer to the left or to the right when approaching a previously unseen obstacle. A policy trained according to MSE would most likely fail in such an event and choose to go straight instead, as the learned policy would choose an action that averages between the two equally good alternatives from the dataset.

To tackle these issues, we propose a different paradigm where we assume the expert policy to be composed of two separate policies: a policy $\pi^{*}_{d}$ for discrete actions and a policy $\pi^{*}_{c}$ for continuous actions. Under this assumption, we propose to learn each policy separately, with different learning paradigms: FF-based BC to learn $\pi_{d}$, responsible for modeling the discrete actions $\mathbf{a}_{d}$, and EBM-based BC to learn $\pi_{c}$, responsible for modeling the continuous actions $\mathbf{a}_{c}$.  Assuming a handheld gaming controller, discrete actions $\mathbf{a}_{d}$ indicate button presses, while continuous actions $\mathbf{a}_{c}$ indicate joysticks movements and trigger presses. 

These learning paradigms represent the building blocks of \proposed{} and are presented in detail in the following subsections.

\subsection{FeedForward-BC (\purebc{})}
In the \purebc{} stream (blue lane in Fig.~\ref{fig:fig2}), we solve the optimisation problem of Eq.~\ref{eq:opt_prob} via supervised learning methods. When approaching BC via FF deep neural networks (DNN) $F_{\theta}(\cdot)$, we can model both continuous control policies with a regressor, and discrete control policies with a classifier. 

For \proposed{}, we use a \bclane{} classifier to model the learned policy $\hat{\pi}_{d}$. The discrete actions $\mathbf{a}_{d} \in \left\lbrace 0,1 \right\rbrace$, i.e., the button presses on the controller, can be modelled and formalised as a set of binary classification problems, i.e., one for each button. Given the dataset $\mathcal{D}$, we define the corresponding optimisation problem as:
\begin{equation}
    \mathcal{P}_{\purebc{}}: \arg \min_{\theta} \sum_{b} \sum_{k} \sum_{t} BCE (F_{\theta} (\mathbf{s}), \mathbf{a}_{d}) _,
    \label{eq:ff_opt}
\end{equation}
where $b, k, t$ indicate batch, trajectory and time-step, respectively, and $BCE(\cdot, \cdot)$ identifies the binary cross entropy loss used to train the model, as
\begin{equation}
    BCE (f(\mathbf{s}), \mathbf{a}) = \mathbf{a} \cdot \log(f(\mathbf{s})) + (1-\mathbf{a})\cdot \log(1-f(\mathbf{s})).
\end{equation}

To solve this problem, we use a simple FF model with Convolutional Neural Network (CNN) layers for feature extraction as detailed below. 

\noindent\textit{\purebc{} Model.} The stream comprises of a pretrained EfficientNet model for RGB feature extraction~\cite{Tan2019}, a simple CNN for depth feature extraction (2-cells deep CNN and 3-cells deep FF projection) and a FF-based network for telemetry feature extraction. The features are then concatenated in a single vector processed through a cascade of three FF-based networks. The  final projection is split into multiple heads, one for each dimension of the discrete actions vector $\mathbf{a}_{d}$.

\subsection{EnergyControlled-BC (EC-BC)}
In the EC-BC stream (purple lane in Fig.~\ref{fig:fig2}), we solve the optimisation problem of Eq.~\ref{eq:opt_prob} via EBMs. In this case, the policy function is a continuous energy function $E_{\theta}(\cdot)$, namely \ebmlane{}, which can be used to solve classification and regression problems. 

In \proposed{}, we use a \ebmlane{} regressor for the learned policy $\hat{\pi}_{c}$, as continuous actions on the controller $\mathbf{a}_{c} \in \left[ 0,1 \right]$, i.e., joystick and trigger movements, can be modelled as a set of bound regression problems. Given the dataset $\mathcal{D}$,  we formalise the EBM-based optimisation problem as
\begin{equation}
    \mathcal{P}_{EC-BC}: \hat{\mathbf{a}}_{c} = \arg \min_{\mathbf{a}_{c} \sim \mathcal{A}_{c}, \theta} E_{\theta} (\mathbf{s}, \mathbf{a}_{c}) _,
    \label{eq:ebm_opt}
\end{equation}
where $E_{\theta}$ identifies the energy function and $\mathcal{A}_{c}$ represents the set of all possible continuous actions $\mathbf{a}_{c}$. 

As shown above, the role of the energy function is very different to that of a feed-forward network $F_{\theta}(\cdot)$ which directly outputs the action $\mathbf{a}$ as a function to a given input $\mathbf{s}$, i.e., $\hat{\mathbf{a}} = F_{\theta}(\mathbf{s})$ as in Eq.~(\ref{eq:ff_opt}). Instead, the energy function $E_{\theta}(\cdot)$ models the probability distribution of all possible target values of $\mathbf{a}$, given the input $\mathbf{s}$. Therefore, at inference time, we need estimate the best configuration of the $(\mathbf{s}, \mathbf{a})$ pairs, i.e., $\hat{\mathbf{a}} = \arg \min_{y} E_{\theta}(\mathbf{s}, \mathbf{a})$ as in Eq.~(\ref{eq:ebm_opt}). 

There are many possible algorithms that can be used for inference with EBMs, as shown in~\cite{Florence2022, Gustafsson2020}. Due to space constraints, in this paper we will focus on two approaches that have shown to be particularly effective and simple to implement, namely Grid-Search and No-Grad, presented in Sect.~\ref{sect:ebm_inf}.

\noindent\textit{\ebmlane{} Model.} Similarly to \purebc{}, the stream comprises of a pretrained EfficientNet model for RGB feature extraction~\cite{Tan2019}, a simple CNN for depth feature extraction (2-cells deep CNN and 3-cells deep FF projection) and a FF-based network for telemetry feature extraction. The features are then concatenated together with the action vector $\mathbf{a}_{c}$ into a single vector, which we process through a cascade of three FF-based networks with a single-dimension input for energy.

In the following section, we provide a detailed description of the training and inferring paradigms used for the proposed EBM-BC component of \proposed{}.

%% file: sections/4_ebms.tex
\section{Energy Controlled BC (EC-BC)}
\label{sect:ebms}
In this section, we provide a detailed description of training and inference procedures of the \ebmlane{} regressor in \proposed{}. 

\subsection{Training}
In line with previous approaches from the literature, we train \ebmlane{} according to the Noise Contrastive Estimation (NCE) loss function~\cite{Florence2022, Gustafsson2020}. With this approach, we first generate fake label examples that we generate from a chosen distribution, then we ask the model to differentiate these from the true samples of the dataset. 

Since our focus resides on the application of EBMs to BC, all our formulations use the same notation as in Sect.~\ref{sect:model}. Note that we omit the subscript $c$ to ease notation. Given the state information $\mathbf{s}$ and the action information $\mathbf{a}$, we define the NCE loss as:
\begin{equation}
\small
    \mathcal{L}_{InfoNCE} = \sum_{k=1}^{K} - \log \left( \frac{e^{-E_{\theta}(\mathbf{s}, \mathbf{a}_{k})}}{ e^{-E_{\theta}(\mathbf{s}, \mathbf{a}_{k})} + \sum_{j=1}^{N_{fake}}e^{-E_{\theta}(\mathbf{s}, \mathbf{a}_{j})}} \right) _,
\label{eq:nce}
\end{equation}
where $N_{fake}$ is used to identify the number of fake examples that we generate for each sample from the dataset.

It is clear from Eq.~(\ref{eq:nce}) that the generation of fake examples is pivotal to the successful training of the model. In fact, when training the model with NCE-based methods, we want the model to learn how to discern true samples from fake ones. However, we do not want the difference between the two to be obvious as the model falls into sub-optimal minima where the state information $\mathbf{s}$ becomes irrelevant for the model. 

Since the \ebmlane{} module is dedicated to learning the continuous actions of a game, we design the distributions from which we sample fake examples to resemble the ones of the actions they are trying to imitate. In gaming, the patterns of continuous actions tend to be very similar. For instance, joysticks operate as truncated normal distributions. On the other hand, triggers operate as Poisson distributions most of the time. Therefore, in \proposed{} we have included a module that generates different fake examples for continuous actions for the joysticks and for the triggers.  Including this simple, yet efficient, estimation allows for more robust and stable training, in addition to ensuring that the model experiences a wider spectrum of actions for which to compute the energy. The pseudocode implementation for the training procedure is presented in Algorithm~\ref{alg:Training}.

\begin{algorithm}[t]
\begin{algorithmic}
\REQUIRE $\mathcal{D}:(\mathbf{s}, \mathbf{a})$: State-action pairs from offline dataset
\REQUIRE $N_{fake}$: Number of fake examples $\mathbf{\hat{a}}$
\FOR {$i: 0\rightarrow N_{batch}$}
\STATE Sample fake examples: $\mathbf{\hat{a}}_i^{N_{fake}\times A} \sim \mathcal{D}_{fake}$
\STATE Concatenate actions: $\mathbf{a}_{i}= \mathbf{a}_{i} \oplus \mathbf{\hat{a}}_i$
\STATE Compute energy of the state-action pairs: $E_{\theta}(\mathbf{\mathbf{s}_{i}, \mathbf{a}_{i}})$
\STATE Compute loss:  $\mathcal{L}_{InfoNCE} (E_{\theta})$

\ENDFOR
\end{algorithmic}
\caption{EBM Epoch training}
\label{alg:Training}
\end{algorithm}

\subsection{Inference}
\label{sect:ebm_inf}
The output of \ebmlane{} is the energy of a state-action $(\mathbf{s}, \mathbf{a})$ pair, with lower energies identifying more likely pairs. If we estimate the energy of multiple state-action pairs and the sample $(\mathbf{s}, \mathbf{a}_{i})$ has the lowest energy, the agent should proceed with the $\mathbf{a}_{i}$ action. Ideally, one could evaluate all possible combinations of actions in the action space and identify the pair corresponding to the lowest energy. However, such an approach is often impractical, especially so in the case of continuous actions. 

Therefore, we need to find algorithms that allow us to identify a state-action pair with reasonable confidence, and without requiring iterations over too many samples. In \proposed{}, we use two algorithms that have shown to be particularly effective in the literature: a) grid search~\cite{Lecun2006} and b) no-gradient~\cite{Florence2022}. 

\subsubsection{Grid-Search inference} Grid-Search is a simple algorithm where we discretise each dimension of $\mathbf{a}$ by generating $N_{pin}$ evenly spaced values. After discretisation, we generate a grid of all possible combinations of actions in the action space $\mathcal{A}$ with cardinality $A= |\mathcal{A}|$. Therefore, at inference time, we proceed to compute the state-action pairs for each element in the grid. Despite its simplicity, this has shown to be a viable approach for applications where the action space is reduced, e.g., when we need to control actions with $2$ or $3$ dimensions. Still, it becomes rapidly prohibitive as the dimensions of the action vector increase as the operation cost scales exponentially.

With this approach, we only need to specify the number of pins $N_{pin}$ with which we want to sample each element of the action vector $\mathbf{a}$. A pseudocode implementation of this approach is given in Algorithm~\ref{alg:Inference_1}.

\subsubsection{No-Grad inference} No-Grad is based on an algorithm introduced in~\cite{Florence2022}. At its core, it is an iterative approach, where we refine the search for the best state-action pair over multiple iterations. We initially sample a set of $N_{infer}$ action vectors $\mathbf{a}$, where each element of the vector is sampled from a truncated uniform distribution. After computing this initial set of action vectors $\mathbf{a}$, we enter the iterative stage where first compute the energy of each of the samples and then normalise them all according to a softmax function. Once we have the normalised value of the energy, we resample the actions with replacement according to their normalised energy. In other words, actions with higher probability (or lower energy) will be picked more often to generate a new set of samples. To ensure we investigate a sufficiently large area of the action space, we add to each of the sample noise and clip the samples to ensure the bounds of each of the action elements are respected. After the new set of samples has been computed, we shrink the standard deviation of the distribution used to generate noise. This operation is performed $N_{iter}$ times, after which the action with the normalised energy is selected. 

With this approach, we only need to specify the number $N_{infer}$ of action vectors we want to sample at each iteration and the number of iterations $N_{iter}$. A pseudocode implementation of this approach is given in Algorithm~\ref{alg:Inference_2}.

%% file: sections/5_results.tex
\section{Results}
\label{sect:results}
In this section, we address the following research questions:
\begin{enumerate}
    \item Can \proposed{} learn to operate autonomously in a game from a dataset of collected human trajectories?
    \item Is \proposed{} able to explore the environment with a varied number of behaviours, despite a reduced dataset?
    \item Can we use \proposed{} to train robust autonomous agents for quality assessment in games?
\end{enumerate}

\begin{algorithm}[t]
\begin{algorithmic}
\REQUIRE $N_{pin}$: Number of pins to use per action dimension

\STATE Build grid samples: $\mathbf{\hat{a}}^{N_{pin}\times A}$
\STATE Concatenate actions: $\mathbf{a}= \mathbf{a} \oplus \mathbf{\hat{a}}$
\STATE Compute energy of the state-action pairs: $E_{\theta}(\mathbf{\mathbf{s}, \mathbf{a}})$
\STATE Pick action with lowest energy: $\mathbf{a}^{*} = \arg\min E_{\theta}(\mathbf{\mathbf{s}, \mathbf{a}}) $

\end{algorithmic}
\caption{Grid-Search Inference}
\label{alg:Inference_1}
\end{algorithm}

\begin{algorithm}[t]
\begin{algorithmic}
\REQUIRE $N_{infer}$: Number of examples per action dimension
\REQUIRE $N_{iter}$: Number of iterations
\REQUIRE $\sigma$: Noise parameter
\REQUIRE $\eta$: Parameter to reduce noise deviation
\REQUIRE $\mathbf{a}^{MAX},\mathbf{a}^{min}$: Max and min value of continuous actions.
\STATE Sample examples: $\mathbf{\hat{a}}^{N_{infer}\times A} \sim \mathcal{D}_{infer}$
\FOR {$n_{iter}: 0\rightarrow N_{iter}$}
\STATE Concatenate actions: $\mathbf{a}= \mathbf{a} \oplus \mathbf{\hat{a}}$
\STATE Compute energy of the state-action pairs: $E_{\theta}(\mathbf{\mathbf{s}, \mathbf{a}})$
\STATE Compute probabilities:  $\hat{P}_{iter}: \left[\text{softmax}(-E_{\theta}(\mathbf{\mathbf{s}, \mathbf{a}}))\right]$

\IF{$n_{iter}<N_{iter}$}
    \STATE Sample fake examples with replacement: $\mathbf{\hat{a}} \sim\hat{P}_{iter}$
    \STATE Add noise to samples: $\mathbf{\hat{a}} = \mathbf{\hat{a}} + \mathcal{N}(0,\sigma)$
    \STATE Clamp actions: $\mathbf{\hat{a}} = \text{clip} (\mathbf{\hat{a}})$ s.t. $\mathbf{\hat{a}} \in \left[\mathbf{a}^{MAX},\mathbf{a}^{min}\right]$
\ENDIF

\ENDFOR
\STATE Pick action with highest probability: $\mathbf{a}^{*} = \arg\max \hat{P}_{iter}(\mathbf{\hat{a}}_i) $
\end{algorithmic}
\caption{No-Grad EBM Inference}
\label{alg:Inference_2}
\end{algorithm}

To do so, we show the results of the proposed \proposed{} model and compare it to a standard BC baseline. When evaluating the performance of \proposed{}, we focus on two separate aspects of the game: playing performance and exploration performance. For playing, we use three separate metrics: the number of kills, the percentage of games with at least a kill and the time the agent was able to drive around the environment without a fatal crash, i.e., when the agent cannot move anymore. For exploration, we use four separate metrics: a qualitative representation of the kernel density estimation (kde) of the environment explored; and three quantitative, i.e., Kullback–Leibler (KL) divergence, cross-correlation (cc) and similarity (sim), between the kde of the policy against the one from the dataset. 

\subsection{Data Information}
\begin{figure}[t]
  \centering
  \subfloat[Player's perspective]{
  \frame{\includegraphics[width=0.95\linewidth]{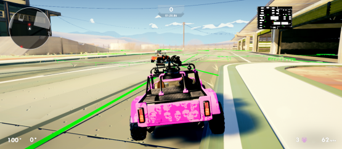}}
  \label{fig:perspective}
  }
  \\
  \centering
  \subfloat[RGB information]{
  \frame{\includegraphics[width=0.45\linewidth]{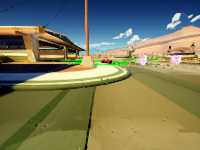}}
  \label{fig:rgb_plot}
  }
  \subfloat[Depth information]{
  \frame{\includegraphics[width=0.45\linewidth]{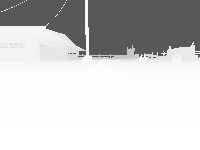}}
  \label{fig:depth_plot}
  }
  \caption{The figures show the (a) interface of the game and the (b)-(c) visual state information collected. }
\label{fig:combo_plot}
\vspace{-0.4cm}
\end{figure}

We trained all policies on a dataset comprising of $60$ demonstrations of $\sim2$ minutes each. Each demonstration corresponds to a full play-through of a death-match level on Hardware Rivals. Hardware Rivals is a driving game, see Fig.~\ref{fig:perspective} for a screenshot of the game, where vehicles are equipped with weapons to damage and destroy the opponents' car. We collect data information from the game at $60~Hz$, for a total of $\sim 432k$ state-action pairs.

Each sample of state information from the game included: a) RGB information from a camera placed on the hood of the vehicle, b) depth information as computed by the game itself, and c) telemetry information. Both RGB and depth information are rendered and captured at a fixed size of $150px\times200px$, to allow for real-time data collection without impacting the performance of the game at run-time. We show an example of RGB and depth data collected in Figs.~\ref{fig:rgb_plot} and~\ref{fig:depth_plot}. The telemetry information comprises of vehicle location on the 2D map $\mathbf{x}: (x_{x}, x_{y})$, direction of the vehicle $\mathbf{d}: (d_{x}, d_{y})$, speed of the vehicle $v$ and ammunition information for the secondary weapon $k$. This information is naturally presented to the player using a head-up display but it is not visible in the RGB information rendered by the camera placed on the hood.

The actions collected during the recording session included: a) navigation inputs (throttle and steering) and the shooting inputs (primary and secondary weapon). Following the same notation as in the previous sections, navigation inputs represented the tuple of the continuous actions $\mathbf{a}_{c}$, whereas shooting inputs were collected as the tuple for the discrete actions $\mathbf{a}_{d}$.

\subsection{Playing Performance}
In this section, we evaluate \proposed{} according to three separate performance metrics: time alive, i.e., the time the agent is able to drive within the environment without severely crashing into obstacles (i.e., to the point of not being able to recover and go back into driving), the average amount of kills it manages to achieve during a match and the percentage of matches where the agents succeeded in getting a kill. We compare the performance of \proposed{} against standard BC approaches and we also provide performance when depth information is not provided. When evaluating both for \proposed{} and standard BC, no hyperparameter tuning was performed and performance was averaged over $20$ matches with a maximum duration of $2$ min.

\begin{figure}[t]
  \centering
  \includegraphics[width=0.99\linewidth,trim=1cm 1cm 1cm 3cm,clip]{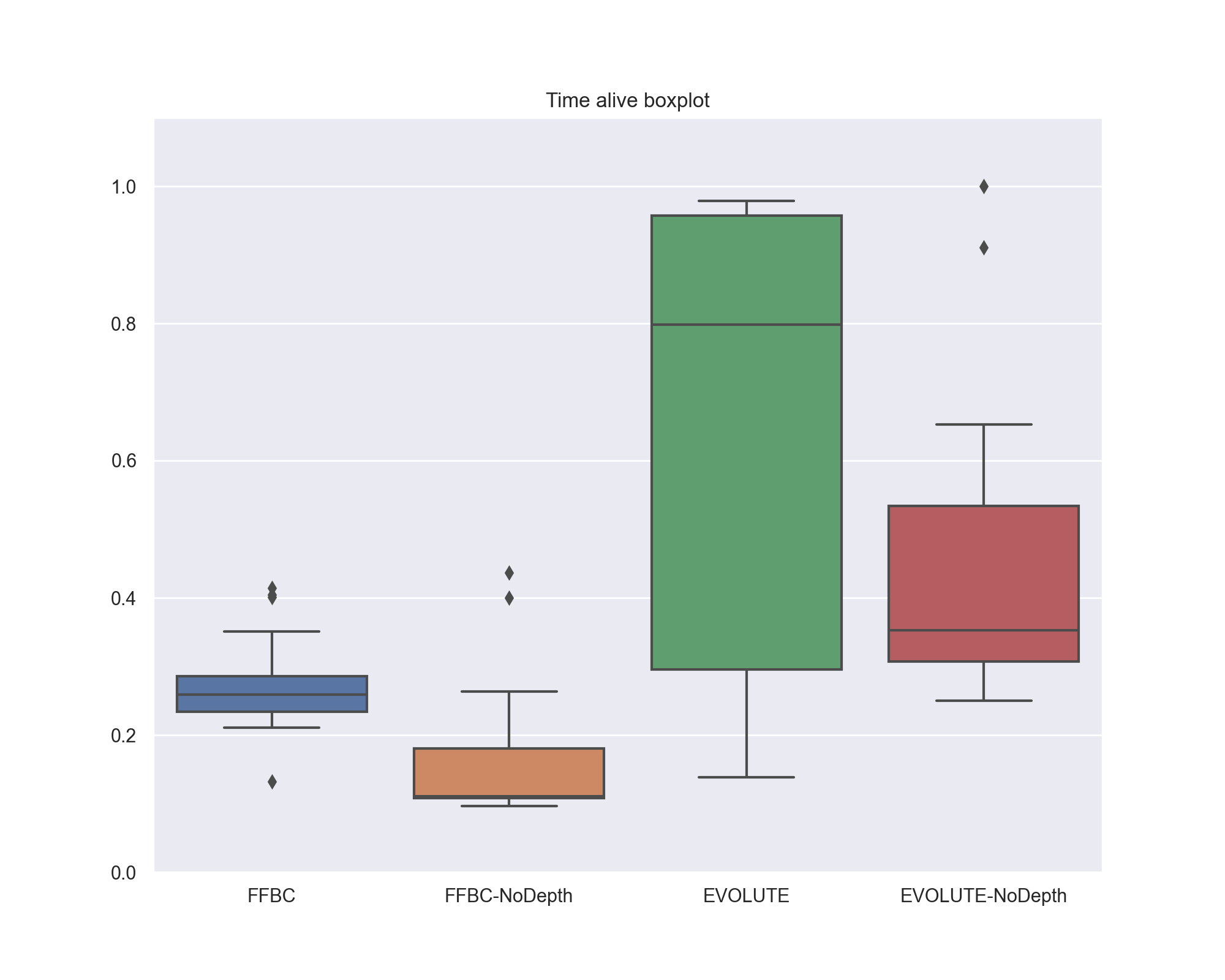}%
  \caption{Time alive. The time alive has been normalised between $[0, 1]$, where $1$ indicates the agent successfully drove within the environment without any critical crash for the duration of a $2$min match. \proposed{} outperforms standard \purebc{}, even when depth is not provided in input.}%
  \label{fig:alive_plot}
\end{figure}

Performance in terms of time alive is collected in Fig.~\ref{fig:alive_plot}. Here, we can see that \proposed{} greatly outperforms \purebc{} in terms of time alive. This is likely due to the probabilistic nature of the \ebmlane{} module in \proposed{}, as recordings showed that standard \purebc{} tends to hit obstacles more often. Due to the MSE loss normally used to train \purebc{}, equally viable options to avoid an obstacle, i.e., steering to the left or steering to the right, often cause the car to simply proceed straight and therefore get stuck. The EBMs loss completely circumvents this problem and provides more robust policies for navigation.

\begin{figure}[t]
  \centering
  \includegraphics[width=0.99\linewidth,trim=1cm 1cm 1cm 3cm,clip]{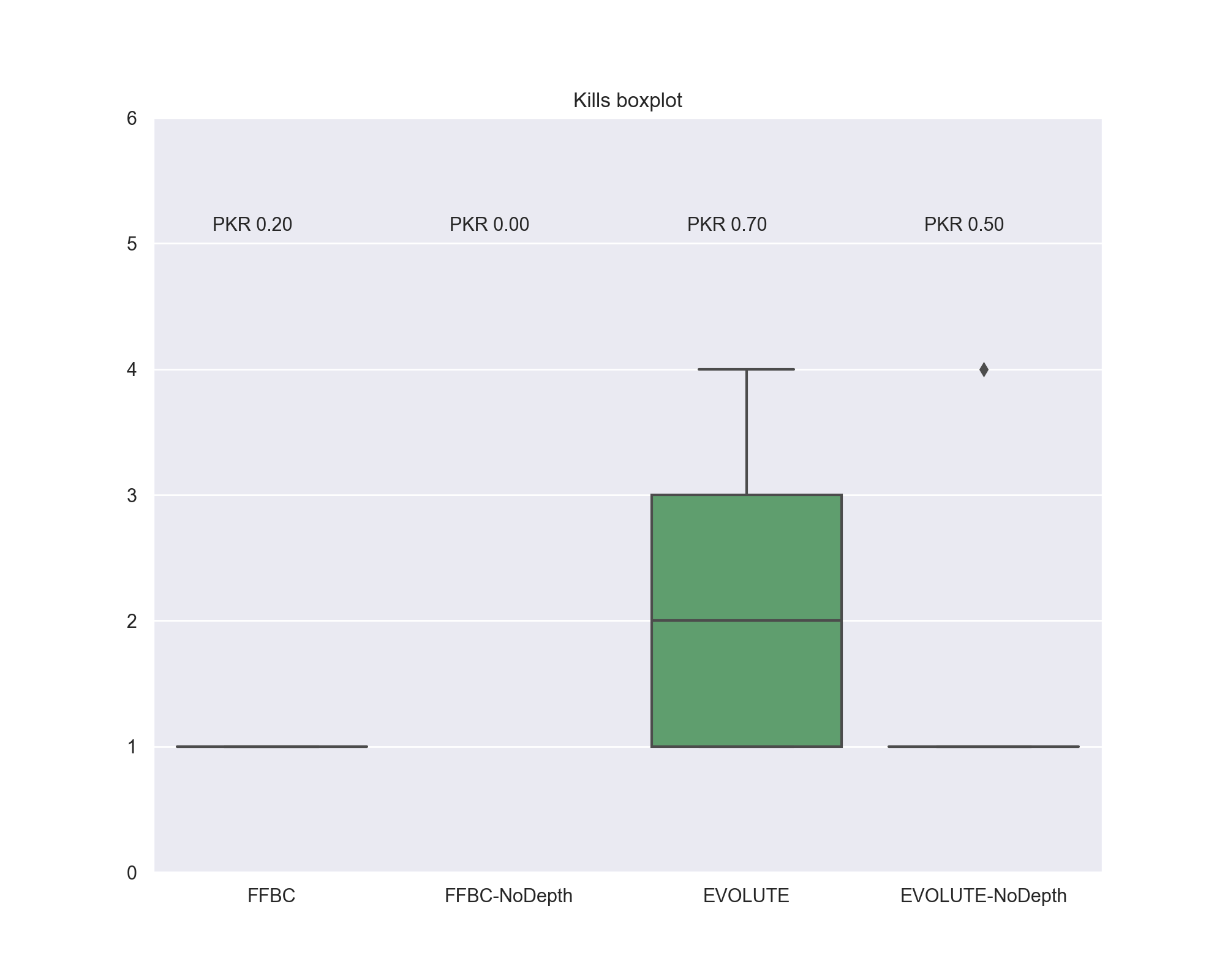}%
  \caption{Kill count. The PKR is computed over $20$ matches, with \proposed{} consistently outperforming standard \purebc{}, even when depth is not provided in input.}%
  \label{fig:kills_plot}
\vspace{-0.4cm}
\end{figure}

Results in Fig.~\ref{fig:kills_plot}, collect the agent's kill performance for both \proposed{} and \purebc{}. In addition to the boxplot of the average kills achieved over 20 matches, we also show the percentage of matches where the agent succeeded in getting a kill, namely a positive kill ratio (PKR). This metric is displayed on the top of the boxplot for all policies. As for the time alive, we can see that \proposed{} greatly outperforms \purebc{} both in terms of the average number of kills it can reach and in terms of PKR. This is not surprising, as \proposed{} can stay alive for longer without hitting obstacles and has therefore more chances to attack enemies and get more kills. On the other hand, since the navigation policy of \purebc{} appears to be more brittle when facing obstacles, it always hits an obstacle before the agent has had a chance to get a kill.

\begin{figure*}[ht]
  \centering
  \subfloat[Dataset location kde ]{
  \includegraphics[width=0.32\linewidth,trim=2cm 1cm 2.5cm 2.25cm,clip]{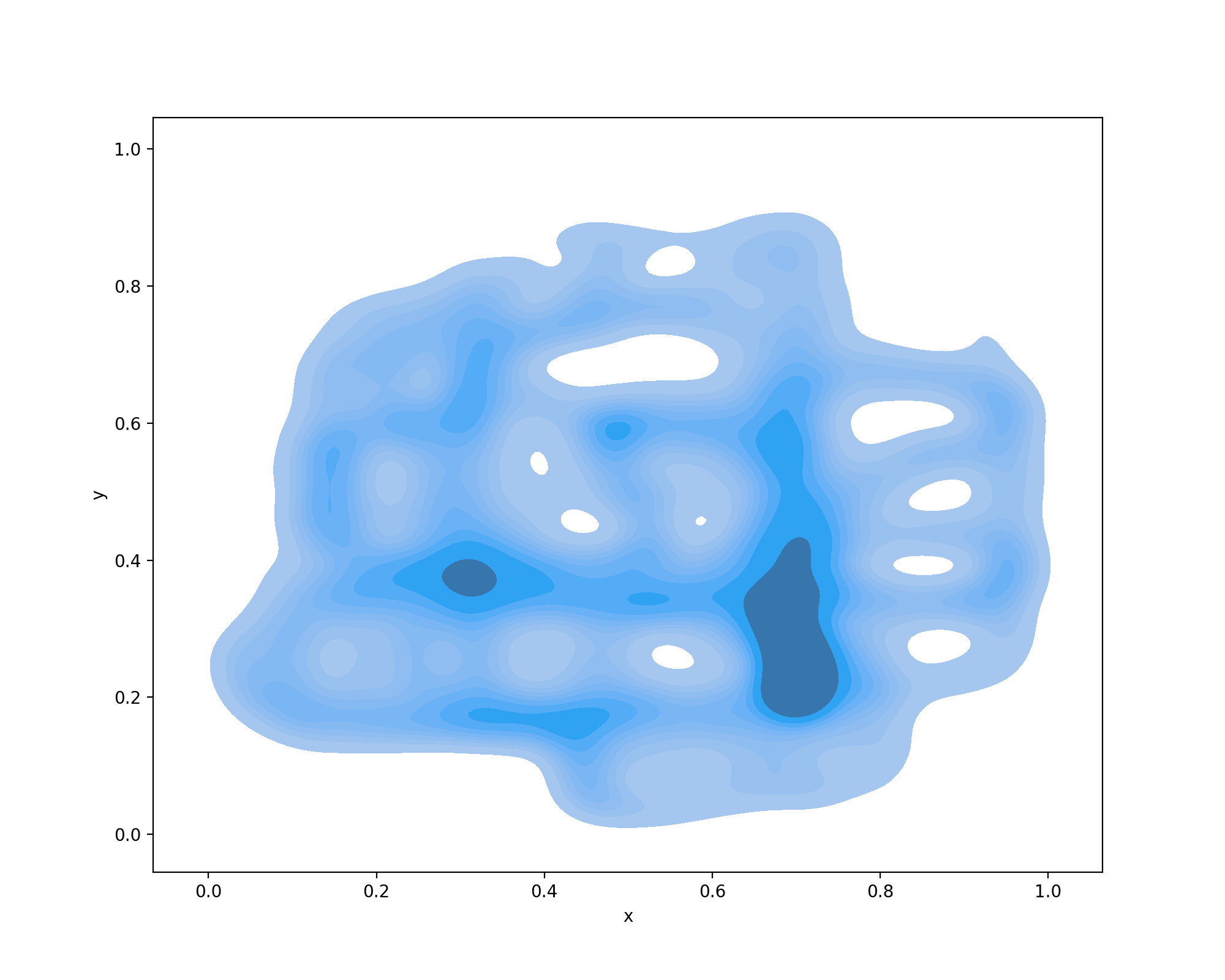}
  \label{fig:kde_data}
  }
  \subfloat[\proposed{} location kde]{
  \includegraphics[width=0.32\linewidth,trim=2cm 1cm 2.5cm 2.25cm,clip]{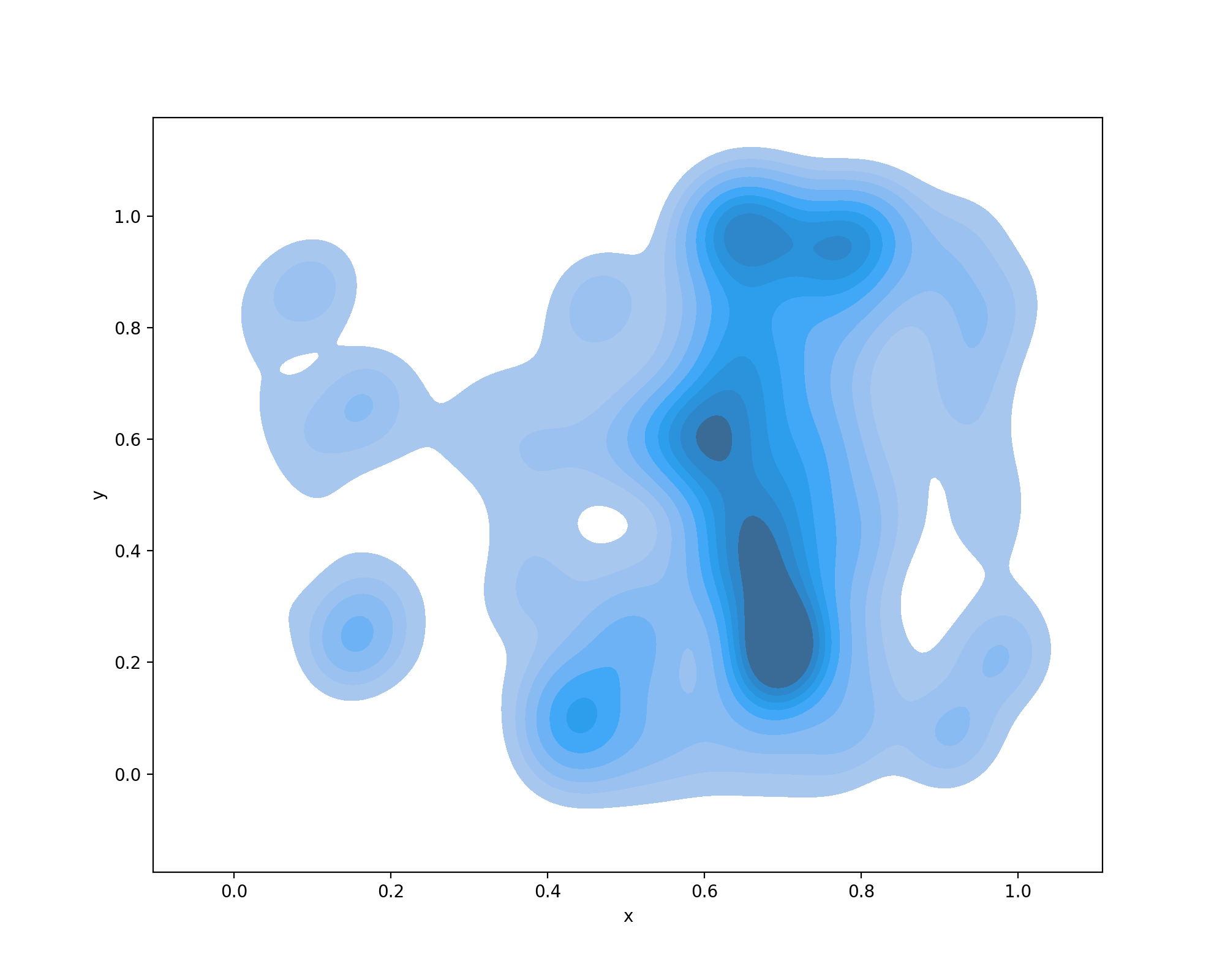}
  \label{fig:kde_evolute}
  }
  \subfloat[\purebc{} location kde]{
  \includegraphics[width=0.32\linewidth,trim=2cm 1cm 2.5cm 2.25cm,clip]{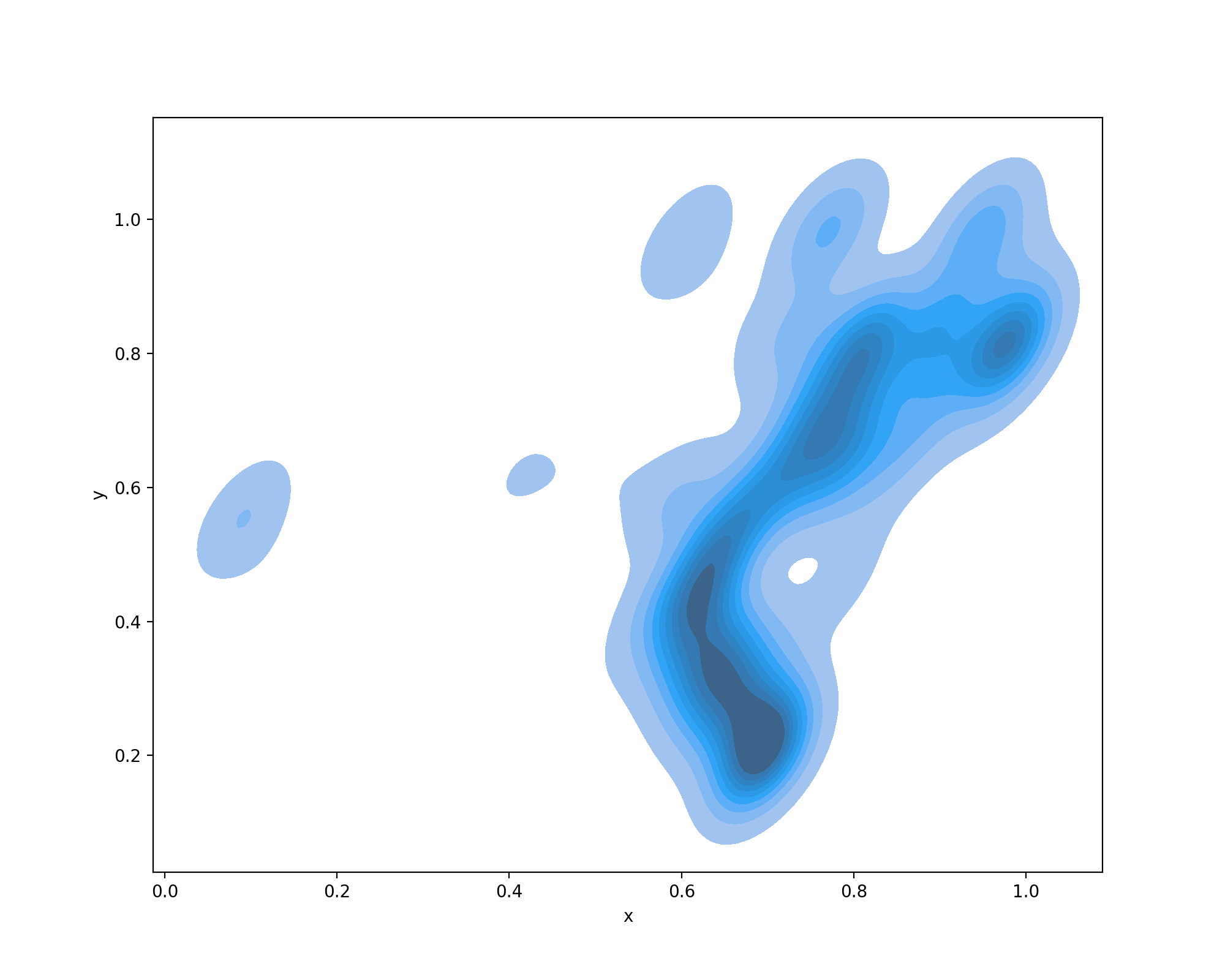}
  \label{fig:kde_ffbc}
  }
  \caption{kde of the $(x,y)$ location data of the expert (a), the \proposed{} policy (b) and the \purebc{} policy (c). }
\label{fig:exploration_kde}
\vspace{-0.3cm}
\end{figure*}

\subsection{Exploration Performance}
In this section, we evaluate the exploration performance for \proposed{} and \purebc{}. When evaluating the exploration performance we collect the $(x, y)$ location of the car when it is being controlled by the trained policy and perform a kde over a collection of 20 runs. For all estimates we use a Gaussian kernel and the estimates are normalised so that they sum to 1. Inspired by the metrics commonly used in visual attention literature~\cite{Amadori2021}, we compare the estimated density of the policy $P$ with the one from the offline dataset $Q^{data}$ and compute the KL divergence, cross-correlation and similarity metrics to quantify their closeness.

The $KL(P,Q) \in \mathbb{R}^{+}$ divergence measures how the probability distribution $P$ is different from a probability distribution $Q$, as
\begin{equation}
KL (P, Q^{data}) = \sum_{i}^{\mathcal{I}} Q^{data}_{i} \log\left(\epsilon + \frac{Q^{data}_{i}}{\epsilon + P_{i}}\right) ,
\end{equation}
where $\sum_{i}$ is the sum over all the bins we used to compute the kde over the sample space $\mathcal{I}$, and $\epsilon$ is the regularizing factor that ensures no division by $0$ occurs and that we never compute the logarithm of $0$. Closer distributions have lower KL values. 

The $CC(P,Q) \in [-1,1]$ metric measures the correlation between two distributions, as
\begin{equation}
CC (P, Q^{data}) = \sum_{i}^{\mathcal{I}} \frac{\text{Cov}\left(P, Q^{data}\right)}{\text{Cov}(P) \times \text{Cov}(Q^{data})} ,
\end{equation}
where the $\text{Cov}(\cdot)$ is the covariance operator. Closer distributions have higher values of $CC$.

The $Sim(P,Q) \in [0,1]$ metric measures the similarity between two distributions, as
\begin{equation}
Sim(P, Q^{data}) = \sum_{i}^{\mathcal{I}} \min(P_{i}, Q^{data}_{i}),
\end{equation}
where $Sim=0$ indicates dissimilar distributions and $Sim=1$ indicates identical distributions. 

As shown in Fig.~\ref{fig:exploration_kde}, \proposed{} is able to explore a wider area of the map, showing a pattern that is consistent with the one of the data it was trained on. While some regions are partially unexplored, see the lower left quadrant in Fig.~\ref{fig:kde_data}, it was still able to reach that area, albeit less frequently than in the original dataset. On the other hand, \purebc{} is really struggling to generalise and explore wider regions of the map, focusing its exploration over the region to the right as visible in Fig.~\ref{fig:kde_ffbc}. This is not very surprising, as previous results on time alive have already shown that \purebc{} often hits obstacles and gets stuck.

To numerically evaluate how closely the policies succeed in exploring the map, we compute KL, CC, and Sim of \purebc{} and \proposed{} and compare the kde of the location they visited with the one from the dataset. The results are collected in Table~\ref{tbl:kde_numerical}. As we can see in the table, \proposed{} is again outperforming \purebc{} with lower scores of KL and higher CC and Sim values. This is consistent with the previous results. Interestingly, we can also see that even \proposed{} without depth information outperforms a full \purebc{}. These results further support the benefits and the generalisation capabilities of \proposed{} for robust testing.

\begin{table}[t]
\setlength{\tabcolsep}{.12cm} 
\centering
\resizebox{.99\linewidth}{!}{
\begin{threeparttable}[t]
\caption{Exploration Performance}
\label{tbl:kde_numerical}
\vspace*{-.1cm}
\renewcommand{\arraystretch}{1.1}
\begin{tabular}{lcccc}
 & \purebc{} & \purebc{}-NoDepth & \proposed{} & \proposed{}-NoDepth
 \\
\hline
{\textit{KL} $\downarrow$} & 7.21 & 7.89 & \underline{2.70} & 4.16 \\
{\textit{CC} $\uparrow$} & -0.03 & -0.04 & \underline{0.35} & 0.22 \\
{\textit{Sim} $\uparrow$} & 0.16 & 0.11 & \underline{0.45} & 0.38 \\
\hline
\end{tabular}
\begin{tablenotes}
    \item[$\uparrow$] Higher values indicate better exploration performance.
    \item[$\downarrow$] Lower values indicate better exploration performance.
\end{tablenotes}
\end{threeparttable}
}
\vspace*{-.4cm}
\end{table}

%% file: sections/6_conclusions.tex
\section{Conclusions}
\label{sect:conclusions}

In this paper, we addressed the problem of automated play testing via a flexible imitation learning framework named \proposed{}. Play testing is a very costly step of game development as human testers are required to repeatedly play many sections of the game to identify possible errors. Even current automated play testing can be very demanding, as it requires sophisticated software engineering for standard tree search models or, in the case of reinforcement learning-based agents, training may require the game to be developed to include specific features for episodic tasks.

We have shown that \proposed{} only requires a limited amount of data to perform a complex task, namely death-match style gameplay, in Hardware Rivals. In addition to performing the task and playing the game, our agent is also able to explore the environment, greatly outperforming standard behavioural cloning. Also, the proposed approach of an ensemble model, where button presses are performed by FF-based behavioural cloning and continuous actions are modelled via EBM-based behavioural cloning, helps reduce the complexities of training EBMs. These results indicate that the proposed model is a viable approach to automate game play testing, while using only a fraction of the human interaction of current approaches.

While the proposed model is able to perform quite well, when compared to human players and better than the other baseline approach, it still inherits some of the limitations of imitation learning. For instance, the proposed approach requires humans to perform the task numerous to provide the data to train the model. While a natural candidate to solve this issue is reinforcement learning, it would still require human intervention on the engine, e.g., for rewards to be computed or for episodic deployment to be possible, among other requirements. On the other hand, the proposed approach only needs access to be able to store the behavioural and game-play data. 

Our results have shown that \proposed{} could be used as a viable approach to automated game testing, thanks to its ability to adapt and explore the environment with limited training data and virtually no hyperparameter tuning. We have also shown that the agent is capable of performing the complex death-match task, succeeding in getting kills and exploring the environment. In the future, it would be interesting to investigate the application of \proposed{} to additional games. Also, the proposed approach makes use of pre-trained models to ease model training and deployment, however we would like to explore fully end-to-end approaches to \proposed{} as they could further improve its performance.

%% file: main.bbl
\begin{thebibliography}{10}
\providecommand{\url}[1]{#1}
\csname url@samestyle\endcsname
\providecommand{\newblock}{\relax}
\providecommand{\bibinfo}[2]{#2}
\providecommand{\BIBentrySTDinterwordspacing}{\spaceskip=0pt\relax}
\providecommand{\BIBentryALTinterwordstretchfactor}{4}
\providecommand{\BIBentryALTinterwordspacing}{\spaceskip=\fontdimen2\font plus
\BIBentryALTinterwordstretchfactor\fontdimen3\font minus \fontdimen4\font\relax}
\providecommand{\BIBforeignlanguage}[2]{{%
\expandafter\ifx\csname l@#1\endcsname\relax
\typeout{** WARNING: IEEEtran.bst: No hyphenation pattern has been}%
\typeout{** loaded for the language `#1'. Using the pattern for}%
\typeout{** the default language instead.}%
\else
\language=\csname l@#1\endcsname
\fi
#2}}
\providecommand{\BIBdecl}{\relax}
\BIBdecl

\bibitem{Politowski2021}
C.~Politowski, F.~Petrillo, and Y.-G. Gu{\'e}h{\'e}neuc, ``A survey of video game testing,'' in \emph{2021 IEEE/ACM International Conference on Automation of Software Test (AST)}.\hskip 1em plus 0.5em minus 0.4em\relax IEEE, 2021, pp. 90--99.

\bibitem{Wurman2022}
P.~R. Wurman, S.~Barrett \emph{et~al.}, ``Outracing champion {Gran Turismo} drivers with deep reinforcement learning,'' \emph{Nature}, vol. 602, no. 7896, pp. 223--228, 2022.

\bibitem{Schaal1999}
S.~Schaal, ``Is imitation learning the route to humanoid robots?'' \emph{Trends in cognitive sciences}, vol.~3, no.~6, pp. 233--242, 1999.

\bibitem{Johns2021}
E.~Johns, ``Coarse-to-fine imitation learning: {R}obot manipulation from a single demonstration,'' in \emph{2021 IEEE international conference on robotics and automation (ICRA)}.\hskip 1em plus 0.5em minus 0.4em\relax IEEE, 2021, pp. 4613--4619.

\bibitem{Mu2021}
T.~Mu, Z.~Ling \emph{et~al.}, ``Maniskill: {G}eneralizable manipulation skill benchmark with large-scale demonstrations,'' \emph{arXiv preprint arXiv:2107.14483}, 2021.

\bibitem{Bojarski2016}
M.~Bojarski, D.~Del~Testa \emph{et~al.}, ``End-to-end learning for self-driving cars,'' \emph{arXiv preprint arXiv:1604.07316}, 2016.

\bibitem{Bansal2018}
M.~Bansal, A.~Krizhevsky, and A.~Ogale, ``{ChauffeurNet}: {L}earning to drive by imitating the best and synthesizing the worst,'' \emph{arXiv preprint arXiv:1812.03079}, 2018.

\bibitem{Pomerleau1991}
D.~A. Pomerleau, ``Efficient training of artificial neural networks for autonomous navigation,'' \emph{Neural computation}, vol.~3, no.~1, pp. 88--97, 1991.

\bibitem{Ross2011}
S.~Ross, G.~Gordon, and D.~Bagnell, ``A reduction of imitation learning and structured prediction to no-regret online learning,'' in \emph{Proceedings of the fourteenth international conference on artificial intelligence and statistics}.\hskip 1em plus 0.5em minus 0.4em\relax JMLR Workshop and Conference Proceedings, 2011, pp. 627--635.

\bibitem{Laskey2017}
M.~Laskey, J.~Lee \emph{et~al.}, ``{DART}: {N}oise injection for robust imitation learning,'' in \emph{Conference on robot learning}.\hskip 1em plus 0.5em minus 0.4em\relax PMLR, 2017, pp. 143--156.

\bibitem{Florence2022}
P.~Florence, C.~Lynch \emph{et~al.}, ``Implicit behavioral cloning,'' in \emph{Conference on Robot Learning}.\hskip 1em plus 0.5em minus 0.4em\relax PMLR, 2022, pp. 158--168.

\bibitem{Lecun2006}
Y.~LeCun, S.~Chopra \emph{et~al.}, ``A tutorial on energy-based learning,'' \emph{Predicting structured data}, vol.~1, no.~0, 2006.

\bibitem{Gustafsson2020}
F.~K. Gustafsson, M.~Danelljan \emph{et~al.}, ``How to train your energy-based model for regression,'' \emph{arXiv preprint arXiv:2005.01698}, 2020.

\bibitem{Grathwohl2019}
W.~Grathwohl, K.-C. Wang \emph{et~al.}, ``Your classifier is secretly an energy based model and you should treat it like one,'' \emph{arXiv preprint arXiv:1912.03263}, 2019.

\bibitem{Zhao2016}
J.~Zhao, M.~Mathieu, and Y.~LeCun, ``Energy-based generative adversarial network,'' \emph{arXiv preprint arXiv:1609.03126}, 2016.

\bibitem{Mnih2013}
V.~Mnih, K.~Kavukcuoglu \emph{et~al.}, ``Playing {Atari} with deep reinforcement learning,'' \emph{arXiv preprint arXiv:1312.5602}, 2013.

\bibitem{Ng2000}
A.~Y. Ng, S.~Russell \emph{et~al.}, ``Algorithms for inverse reinforcement learning,'' \emph{PMLR}, pp. 663--670, 2000.

\bibitem{Ho2016}
J.~Ho and S.~Ermon, ``Generative adversarial imitation learning,'' \emph{Advances in neural information processing systems}, vol.~29, 2016.

\bibitem{Wang2019}
R.~Wang, C.~Ciliberto \emph{et~al.}, ``Random expert distillation: {Imitation} learning via expert policy support estimation,'' \emph{PMLR}, vol.~97, pp. 6536--6544, 2019.

\bibitem{Silver2017}
D.~Silver, J.~Schrittwieser \emph{et~al.}, ``Mastering the game of go without human knowledge,'' \emph{nature}, vol. 550, no. 7676, pp. 354--359, 2017.

\bibitem{Vinyals2019}
O.~Vinyals, I.~Babuschkin \emph{et~al.}, ``Grandmaster level in {StarCraft II} using multi-agent reinforcement learning,'' \emph{Nature}, vol. 575, no. 7782, pp. 350--354, 2019.

\bibitem{Ariyurek2019}
S.~Ariyurek, A.~Betin-Can, and E.~Surer, ``Automated video game testing using synthetic and humanlike agents,'' \emph{IEEE Transactions on Games}, vol.~13, no.~1, pp. 50--67, 2019.

\bibitem{Lillicrap2015}
T.~P. Lillicrap, J.~J. Hunt \emph{et~al.}, ``Continuous control with deep reinforcement learning,'' \emph{arXiv preprint arXiv:1509.02971}, 2015.

\bibitem{Cai2023}
M.~Cai, E.~Aasi \emph{et~al.}, ``Overcoming exploration: {Deep} reinforcement learning for continuous control in cluttered environments from temporal logic specifications,'' \emph{IEEE Robotics and Automation Letters}, vol.~8, no.~4, pp. 2158--2165, 2023.

\bibitem{Delalleau2019}
O.~Delalleau, M.~Peter \emph{et~al.}, ``Discrete and continuous action representation for practical rl in video games,'' \emph{arXiv preprint arXiv:1912.11077}, 2019.

\bibitem{Tan2019}
M.~Tan and Q.~Le, ``Efficientnet: {Rethinking} model scaling for convolutional neural networks,'' in \emph{International conference on machine learning}.\hskip 1em plus 0.5em minus 0.4em\relax PMLR, 2019, pp. 6105--6114.

\bibitem{Amadori2021}
P.~V. Amadori, T.~Fischer, and Y.~Demiris, ``{HammerDrive: A} task-aware driving visual attention model,'' \emph{IEEE Transactions on Intelligent Transportation Systems}, vol.~23, no.~6, pp. 5573--5585, 2021.

\end{thebibliography}
